\documentclass[envcountreset]{llncs}
\usepackage{makeidx}  % allows for indexgeneration
\usepackage{times}
\usepackage{helvet}
\usepackage{courier}
\usepackage{multirow}
\usepackage{amsfonts}
\usepackage{amsmath}
\usepackage{bm}
\usepackage{algorithmic}
\usepackage{graphicx}
\usepackage{subfigure}
\usepackage{epstopdf}
\usepackage[ruled,vlined]{algorithm2e}

\begin{document}
%
% \title{Unsupervised Feature Analysis with \\ Class Margin Optimization}
\title{Uncovering Locally Discriminative Structure for Feature Analysis\thanks{Corresponding Author: Feiping Nie. Email: feipingnie@gmail.com}}
%
%\titlerunning{Unsupervised Featue Analysis}  % abbreviated title (for running head)
\author{
Sen Wang\inst{1}\and
Feiping Nie\inst{2} \and
Xiaojun Chang\inst{3} \\
Xue Li\inst{1} \and
Quan Z. Sheng\inst{5} \and
Lina Yao\inst{4}
}
\institute
{
School of Information Technology and Electrical Engineering,\\ The University of Queensland, Australia. \and
Centre for OPTIMAL, Northwestern Polytechnical University, Shaanxi, China. \and
Centre for Quantum Computation \& Intelligent Systems,\\ University of Technology Sydney, Australia. \and
School of Computer Science, The University of Adelaide, Australia. \and
School of Computer Science and Engineering, The University of New South Wales, Australia. 
}

\maketitle              % typeset the title of the contribution

\begin{abstract}
Manifold structure learning is often used to exploit geometric information among data in semi-supervised feature learning algorithms. In this paper, we find that local discriminative information is also of importance for semi-supervised feature learning. We propose a method that utilizes both the manifold structure of data and local discriminant information. Specifically, we define a local clique for each data point. The k-Nearest Neighbors (kNN) is used to determine the structural information within each clique. We then employ a variant of Fisher criterion model to each clique for local discriminant evaluation and sum all cliques as global integration into the framework. In this way, local discriminant information is embedded. Labels are also utilized to minimize distances between data from the same class. In addition, we use the kernel method to extend our proposed model and facilitate feature learning in a high-dimensional space after feature mapping. Experimental results show that our method is superior to all other compared methods over a number of datasets. 
% \keywords{unsupervised feature selection, maximum margin criterion, sparse structure learning, embedded K-means clustering}
\end{abstract}
%

%\input{introduction}
%!TEX root = main.tex
\section{Introduction}
\label{introduction}
The performance of machine learning tasks, e.g. classification or clustering, is mainly affected by the input features that are extracted from raw data. Learning distinctive features or effective data representations can without doubt benefit the consequent learning tasks. Over the past decade, feature analysis has attracted much research attention in different fields, such as machine learning \cite{wang2015unsupervised,zhu2016robust,wang2016compact,ZhangNX10}, multimedia analysis \cite{wang2016compact}, biomedical applications \cite{zhu2015novel,ZhuLZ16,ZhuSLS16}, etc. In literature, a number of unsupervised and supervised methods have been developed to learn new features. Clustering algorithms have been widely used as an unsupervised feature learning procedure to obtain statistical data representation. A typical example in multimedia analysis is k-means, which learns a dictionary from a number of image or video training samples without class information \cite{yang2009linear,lu2013learning}. Even though increasing the size of the dictionary can squeeze out a bit of extra performance, it is still difficult to identify the best choice of the number of centers.

As one of the representative supervised algorithms, Linear Discriminant Analysis (LDA) \cite{fukunaga1990introduction} finds the best data projection maximizing distances between different class centers while making data samples from the same class closer to each other. This is achieved by maximizing the ratio of the between-class covariance to the within-class covariance. When there are sufficient labeled training data, LDA-based features effectively support machine learning algorithms in a variety of applications. For example, LDA has been used to strengthen the class information of face images in many face recognition systems \cite{goudail1996face,swets1996using,belhumeur1997eigenfaces,ye2004lda,shu2012efficient,ChangNWYZZ16,NieZTXZ11,NieXLHZ12}. Also, LDA has been revisited in \cite{bekios2011revisiting} and been evaluated in three pipelines over a few face image datasets for the purpose of gender recognition. In \cite{iosifidis2012multi}, authors evaluate three LDA-based variants to obtain a discriminant movement representation for multi-view action recognition. Unfortunately, when the number of training samples is small, LDA suffers from the \textit{small sample size} (SSS) problem. This is because the small number of training samples will make the within-class scatter matrix singular, which would result in computational difficulty. Meanwhile, learning new features from a small number of labeled training samples in a fully supervised manner may lead to the over-fitting problem. To solve these problems, much research attention has been paid over the last few years. For example, subspace learning methods, such as Principal Component Analysis (PCA), are applied to reduce feature dimensionality prior to LDA, with the goal of removing null space of the within-class scatter matrix. However, this preprocessing step may lose discriminant information which means the consequent projection in the subspace by LDA may not be the best. A number of methods \cite{chen2000new,yu2001adirect,huang2002solving,kyperountas2007weighted,jiang2008eigenfeature,YangMNCH15,zhang2010generalized} have been proposed to tackle the SSS problem without losing discriminant information. Though the SSS problem is dealt with, the over-fitting problem persists. Increasing the number of labeled training samples would be an ideal solution. However, data labeling in the real world is usually time-consuming and expensive. Considering the huge amount of data without labels, it is extremely difficult to obtain massive and comparable label information. For this reason, semi-supervised feature analysis methods \cite{he2004locality,cai2007semi,li2013low,ChangNYH14}, which make use of both labeled and unlabeled data, have been extensively studied in the past. Unfortunately, most of the existing semi-supervised feature learning algorithms ignore the utilization of both manifold structure and local discriminant information. 

In this paper, we propose a new kernel-based feature learning method that can learn new features when the labeled information is limited. Some researchers \cite{roweis2000nonlinear,yang2008harmonizing,yang2012tpami} have pointed out that exploiting the local structures is more effective and efficient than learning the global structures. Besides, the manifold structure of data is another crucial property to be considered in feature learning. In this work, the contribution can be briefly summarized as the utilization of both the manifold structure and the local discriminant information to deal with the shortage of labeled data points. Compared with those representative semi-supervised feature learning methods, our proposed method not only makes labeled data within the same class closer to each other, but also incorporates the local discriminant information into a joint framework. We find that the local discriminative information of the manifold structure is very important for feature learning, especially when the label information is scarce. This is omitted in most of the previous works on semi-supervised feature learning. In order to exploit the manifold structure and the local discriminant information, we learn a new graph Laplacian. Specifically, for each data point, we define a local clique in which the data point and its $k-1$ geometric neighbors are included. To achieve this point, kNN is used to exploit the intrinsic manifold structure of data. Moreover, we employ a variant of the Fisher criterion to each clique to evaluate the local discriminant information. The sum of all cliques will be integrated into a joint framework as a global integration. In this way, a new graph Laplacian that holds manifold information with local discriminant information can be learned. This method has two-fold advantages: Firstly, since there are only $k$ data points in each clique (normally quite smaller than the dimensionality of data points), the overall computational burden is greatly relieved. This is because calculating an inverse of a $k\times k$ matrix $n$ times is faster than a direct inverse calculation on a $n\times n$ matrix when $n$ is big. Secondly, it is easy to extend the local discriminant model to a kernel-based version. In this work, we also extend our proposed model using kernel method to learn both labeled and unlabeled data in a high-dimensional space in which data are linearly separable. Additionally, we have proposed an algorithm to solve the optimization problem. 

The rest of paper is organized as follows: Notations and definitions will be presented in Section 2. Our proposed method and the optimization algorithm will be elaborated upon in Section 3. In Section 4, we will report the experimental settings, results, and related analysis. The conclusion will be given in the last section.
\section{Notations and Definitions}
\label{pre_pro}
To give a better understanding of the proposed algorithm, notations and definitions used in this paper are summarized in this section. Matrices and vectors are written as boldface uppercase letters and boldface lowercase letters, respectively. In this paper, we follow the conventional definition in semi-supervised learning. In the training dataset, there are $n$ data samples including $m$ labeled data and $n-m$ unlabeled data. Thus, the training data matrix is defined as:
\begin{equation}
\bm{X} = \left[ \begin{array}{cc}
\bm{X}_l&\bm{X}_u \end{array}
\right]
\end{equation}
$\bm{X}\in\mathbb{R}^{d \times n}$. The labeled data are denoted by $\bm{X}_l\in\mathbb{R}^{d \times m}$, and the unlabeled data are denoted by $\bm{X}_u\in\mathbb{R}^{d \times (n-m)}$. $d$ is feature dimensionality. Correspondingly, label assignment matrix $\bm{Y}\in\mathbb{R}^{n \times c}$ is defined as:
\begin{equation}
\bm{Y} = \left[ \begin{array}{c}
\bm{Y}_l\\
\bm{Y}_u \end{array}
\right]
\end{equation}
$\bm{Y}_l \in \mathbb{R}^{m \times c}$ is for the labeled data while $\bm{Y}_u \in \mathbb{R}^{(n-m) \times c}$ is for the unlabeled data. $c$ is the number of classes. $\bm{Y}_u$ is initialized with all zeros and its entry, $y_{it}\in[0,1]$, denotes how likely the $i$-th unlabeled data belongs to the $t$-th class, where $m+1\leq i \leq n$ and $1\leq t \leq c$.

\section{Locally Discriminative Structure Uncovering}
\label{proposedmethod}
\subsection{Proposed method}
%Before elaborating our proposed method in this section, we give notation and definitions that are used through out this paper.
Inspired by kernel methods, we assume that, after a non-linear mapping function $\bm{f}=\bm{\phi}(\bm{x})$, the mapped data within the same class are still geometrically close to each other in a high-dimensional space that $\bm{\phi}(\bm{x})$ maps to. The problem can be formulated as:
\begin{equation}
  \min\limits_{\phi} \sum\limits_{t=1}^c \sum\limits_{\bm{x}_p,\bm{x}_q \in \pi_t}||\bm{\phi}(\bm{x}_p)-\bm{\phi}(\bm{x}_q)||^2
  \label{ori_obj}
\end{equation}
$\pi_t$ contains all the data points from the $t$-th class.
%According to the Representer Theorem \cite{Scholkopf:2001:LKS:559923}, we can have:
Without loss of generality, we can have:
\begin{equation}
  \bm{\phi}(\bm{x})=\sum\limits_{i=1}^m \alpha_i\bm{k}(\bm{x},\bm{x}_i),
  \label{def_kernel}
\end{equation}
$\bm{k}$ is a kernel defined on $\bm{x}_1, \ldots, \bm{x}_m \in {\mathcal{X}}$. Note that Eq. \eqref{ori_obj} needs class information, thus only $m$ labeled data points are considered in Eq. \eqref{ori_obj}.
After substituting \eqref{def_kernel} into \eqref{ori_obj} by a matrix representation, the objective becomes: 
\begin{equation}
  \begin{aligned}
	  \min\limits_{\bm{\alpha}} &\sum\limits_{t=1}^c \sum\limits_{\bm{x}_p,\bm{x}_q \in \pi_t}|| \bm{\alpha}^T \bm{k}(\bm{X}^T_l,\bm{x}_p)-\bm{\alpha}^T\bm{k}(\bm{X}^T_l,\bm{x}_q)||^2 \\
	  =\min\limits_{\bm{\alpha}} &\bm{\alpha}^T\bm{K}_l\bm{L}_w\bm{K}_l\bm{\alpha},
  \label{obj_2}
\end{aligned}
\end{equation}
where
\begin{equation}
	\bm{L}_w=\bm{D}-\bm{W}
  \label{Lw}
\end{equation}
$\bm{\alpha}=[\alpha_1, \ldots, \alpha_m]^T \in \mathbb{R}^m$ is a vector. 
The weight matrix $\bm{W}_l\in\mathbb{R}^{m\times m}$ for labeled data is defined as:
\begin{equation}
	\bm{W}_{ij}= \begin{cases}
    1 & x_i \text{ and } x_j \text{ are in the same class}; \\
    0 & \text{otherwise,}
  \end{cases}
  \label{W}
\end{equation}
$\bm{D}$ is a diagonal matrix with $\bm{D}_{ii}=\sum_{j=1}^m \bm{W}_{ij}$. $\bm{K}_l\in\mathbb{R}^{m\times m}$ is the kernel matrix of labeled samples. Note that the representation of $\bm\alpha$  in \eqref{obj_2} is an $m$-dimensional vector that transforms the original feature into one dimension. Now, we extend the learned feature into $r$ dimensions by using a transformation matrix $\bm a = [\bm \alpha_1, \ldots, \bm \alpha_r]\in\mathbb{R}^{m\times r}$. Consequently, the objective function in \eqref{obj_2} is equivalent to
\begin{equation}
	\min\limits_{\bm{a}^T \bm{K}_l \bm{a}=I}Tr(\bm{a}^T \bm{K}_l\bm{L}_w\bm{K}_l\bm{a}),
  \label{obj_3}
\end{equation}
$Tr(\cdot)$ is trace operator. The constraint $\bm{a}^T \bm{K}_l \bm{a}=I$ is to make the projection orthogonal. 

The objective in \eqref{obj_3} is based on label information. We aim to make use of both labeled and unlabeled data to improve the performance when the class information is in scarcity. We now extend it into a semi-supervised method by extending the weighted matrix $\bm{W}_l$ in \eqref{W} into a new one:
\begin{equation}
	\bm{W} = \left[\begin{array}{ll}
			\bm{W}_l&\bm{0}_{m \times (n-m)}\\
			\bm{0}_{(n-m) \times m}&\bm{0}_{(n-m) \times (n-m)}
\end{array}
\right]
\end{equation}
$\bm{W}_l$ is the weighted matrix for labeled data. $\bm{0}_{m\times (n-m)}$ is a zero matrix with $m$ rows and $(n-m)$ columns. In this way, the adjacency graph $\bm{L}_w$ is enlarged from $m\times m$ to $n\times n$. 

As mentioned before, we find that local discriminant information among data, labeled and unlabeled, is quite useful for learning new features especially when the label information is limited. In this work, we aim to learn such local structures and to embed the structures into a joint framework. The objective function in \eqref{obj_3} can be re-written as:
\begin{equation}
	\min\limits_{\bm{a}^T \bm{K} \bm{a}=I}Tr(\bm{a}^T \bm{K}\bm{L}_w\bm{K}\bm{a}) + \lambda \Omega(\cdot),
  \label{obj_31}
\end{equation}
$\Omega(\cdot)$ is the regularization term that exploits the local structures among data samples. $\lambda$ is the regularization parameter. We assume that all data within the same class are put together and define a scaled class assignment matrix $\bm{G}\in\mathbb{R}^{n\times c}$ as follows:
\begin{equation}
	\bm{G}=\bm{Y}(\bm{Y}^T\bm{Y})^{-\frac{1}{2}}
\end{equation}
$\bm{Y}=[\bm{y_1}, \ldots, \bm{y_n}]^T\in\mathbb{R}^{n\times c}$. To exploit the local structure, we define a clique of $\bm{x}_i$, denoted as $\bm{\mathcal{N}}_k(\bm{x}_i)$, which has $k$ data samples containing $\bm{x}_i$ itself and its $k-1$ neighbors. Given the data matrix $\bm{X}\in\mathbb{R}^{d\times n}$, the local data matrix for the $i$-th data is defined as $\bm{X}_i=[\bm{x}_i,\bm{x}_{i_1},\ldots,\bm{x}_{i_{k-1}}]\in\mathbb{R}^{d\times k}$.
%, $\bm{x}_{i_s}\in\bm{\mathcal{N}}_k(\bm{x}_i)$, $1\le s\le k-1$. 
Correspondingly, the local scaled classification matrix for $\bm{x}_i$ can be defined as $\bm{G}_{i}=[\bm{g}_i, \ldots, \bm{g}_{i_{k-1}}]^T \in\mathbb{R}^{k\times c}$. Note that both $\bm{X}_i$ and $\bm{G}_i$ are actually selected from $\bm{X}$ and $\bm{G}$ respectively. We define a selection matrix $\bm{S}_i\in\mathbb{R}^{n\times k}$ for $\bm{x}_i$ where $\bm{S}_i^{pq}=1$, if $\bm{x}_p$ is the $q$-th element of $\bm{\mathcal{N}}_k(\bm{x}_i)$, $\bm{S}_i^{pq}=0$ otherwise. $1\le q\le k$. Thus, the local scaled classification matrix for each data can be re-written as $\bm{G}_i=\bm{S}_i^T\bm{G}=\bm{S}_i^T\bm{Y}(\bm{Y}^T\bm{Y})^{-\frac{1}{2}}$.

In linear discriminant analysis, the objective is to maximize the separability of all data points and to minimize the distances among data that are from the same class. According to the definitions of scatter matrix in linear discriminant analysis, the corresponding total scatter and between class scatter matrices for the local data clique can be defined as $\bm{S}_{t_i}$ and $\bm{S}_{b_i}$. To simplify the formulation, we centralize each local data matrix $\bar{\bm{X}_i} = \bm{X}_i\bm{H}$, where $\bm{H}=\bm{I}_{k\times k}-\frac{1}{k}\bm{1}_k\bm{1}_k^T$. $\bm{1}_k$ is a $k$-dimensional column vector with all ones. Thus, we have:
\begin{equation}
\begin{aligned}
\bm{S}_{t_i}&= \sum\limits_{i=1}^n(\bm{x}_i - \bm{\mu})(\bm{x}_i - \bm{\mu})^T = \bar{\bm{X}_i}\bar{\bm{X}_i}^T \\
\bm{S}_{b_i}&= \sum\limits_{t=1}^c n_t(\bm{\mu}_t - \bm{\mu})(\bm{\mu}_t - \bm{\mu})^T = \bar{\bm{X}_i}\bm{G}_i \bm{G}_i^T\bar{\bm{X}_i}^T
\label{obj_32}
\end{aligned}
\end{equation}
where $\bm{\mu}_t$ is the mean of samples in the $t$-th class and $\bm{\mu}$ is the global mean that is zero after the centralization. $n_t$ is the number of data points of the $t$-th class. Inspired by the Fisher criterion \cite{fukunaga1990introduction}, the optimal local scaled class assignment matrix $\bm{G}_i^*$ can be obtained by optimizing the follow objective:
\begin{equation}
\begin{aligned}
\label{obj_4}
\bm{G}_i^*&=\arg\max\limits_{\bm{G}_i} Tr\left((\bm{S}_{t_i} + \theta \bm{I})^{-1}\bm{S}_{b_i}\right) \\
&=\arg\max\limits_{\bm{G}_i} Tr\left(\bm{G}_i^T\bar{\bm{X}_i}^T(\bar{\bm{X}_i}\bar{\bm{X}_i}^T + \theta \bm{I})^{-1}\bar{\bm{
X}_i}\bm{G}_i \right),
\end{aligned}
\end{equation} 
where $\theta \bm{I} (\theta>0)$ is added to avoid $(\bm{S}_{t_i}+ \theta \bm{I})$ be singular. In Eq. \eqref{obj_4}, $\bm{G}_i^*$ is a score that evaluates the local discriminant information of each data points. A larger value indicates that the samples in the local clique from different classes are better separated. To control the capacity of local discriminant model, we add a regularization term $Tr(\bm{G}_i^T\bm{H}\bm{G}_i)$. Then the optimization problem in \eqref{obj_4} is equivalent to
\begin{equation}
\label{obj_5}
\arg\min\limits_{\bm{G}_i} Tr\left(\bm{G}_i^T\bm{H}\bm{G}_i-\bm{G}_i^T\bar{\bm{X}_i}^T(\bar{\bm{X}_i}\bar{\bm{X}_i}^T + \theta \bm{I})^{-1}\bar{\bm{X}_i}\bm{G}_i\right)
\end{equation} 
It is proved in the supplemental document, the problem of \eqref{obj_5} is equivalent to 
\begin{equation}
  \label{obj_6}
  \begin{aligned}
   &\arg\min\limits_{\bm{G}_i} Tr\left(\bm{G}_i^T \bm{L}_i \bm{G}_i \right), \\
   &\text{where } \bm{L}_i = \bm{H}(\bar{\bm{X}_i}^T\bar{\bm{X}_i}+\theta \bm{I})^{-1}\bm{H}
  \end{aligned}
\end{equation}  

Because of $\bm{G}_i=\bm{S}_i^T \bm{G}$, we take all local manifold structures into account together by summing \eqref{obj_6} over all local cliques. Then, the global local discriminant score can be written as:
\begin{equation}
\label{obj_7}
\begin{aligned}
	\arg\min\limits_{\bm{G}} \sum\limits_{i=1}^n Tr(\bm{G}_i^T \bm{L}_i \bm{G}_i)=\arg\min\limits_{\bm{G}} Tr(\bm{G}^T\bm{L}\bm{G}), \\
\end{aligned}
\end{equation}
%=&\arg\min\limits_{\bm{G}} \sum\limits_{i=1}^n Tr(\bm{G}^T\bm{S}_i\bm{L}_i\bm{S}_i^T\bm{G}) \\
%	=&\arg\min\limits_{\bm{G}} Tr(\bm{G}^T\left(\sum\limits_{i=1}^n \bm{S}_i\bm{L}_i\bm{S}_i^T\right)\bm{G}) \\
where 
\begin{equation}
\begin{aligned}
\label{LRGA_L}
\bm{L} &= \sum\limits_{i=1}^n \bm{S}_i\bm{L}_i\bm{S}_i^T \\
\end{aligned}
\end{equation}

By using the graph Laplacian in \eqref{LRGA_L}, local discriminant manifold structures among data points can be therefore embedded into a joint framework. The objective function in \eqref{obj_31} arrives at:
\begin{equation}
\begin{aligned}
	&\min\limits_{\bm{a}^T \bm{K} \bm{a}=\bm{I}}Tr(\bm{a}^T \bm{KL}_w\bm{K}\bm{a}) + \lambda Tr(\bm{a}^T \bm{KLK}\bm{a})\\
	=&\min\limits_{\bm{a}^T \bm{K} \bm{a}=I}Tr(\bm{a}^T \bm{K}(\bm{L}_w+\lambda \bm{L})\bm{K}\bm{a})
  \label{obj_8}
  \end{aligned}
\end{equation}
where $\lambda$ is a parameter that leverages the proportion of utilization of both manifold structure and local discriminant information in the joint framework. 

\subsection{Optimization}
To solve the objective function in \eqref{obj_8}, we assume
\begin{equation}
	\bm{K}=\bm{V\Lambda V}^T,
  \label{K}
\end{equation}
and $\tilde{\bm{V}}$ is the null space of $\bm{V}$. For any $\bm a$, we can represent 
\begin{equation}
	\bm a=\bm{V\beta}+\tilde{\bm{V}}\bm\gamma,
  \label{alpha}
\end{equation}
Thus, 
\begin{equation}
\begin{aligned}
	\bm a^T \bm{K a} & = (\bm{V\beta}+\tilde{\bm{V}}\bm\gamma)^T \bm{K} (\bm{V\beta}+\tilde{\bm{V}}\bm{\gamma}) \\
	&=\bm\beta^T\bm{V}^T\bm{KV\beta}=\bm\beta^T \bm\Lambda \bm\beta,
\end{aligned}
  \label{aka}
\end{equation}
Substituting \eqref{alpha} and \eqref{aka} into \eqref{obj_8}, the objective function is re-formulated as
\begin{equation}
\begin{aligned}
	&\min\limits_{\bm{\beta}^T \bm{\Lambda} \bm{\beta}=\bm{I}}Tr(\bm a^T\bm{K}(\bm{L}_w + \lambda \bm{L}) \bm{K}\bm a)\\
	= &\min\limits_{\bm{\beta}^T \bm{\Lambda} \bm{\beta}=\bm{I}}Tr(\bm\beta^T\bm{\Lambda} \bm{V}^T(\bm{L}_w + \lambda \bm{L}) \bm{V}\bm{\Lambda}\bm\beta),
\end{aligned}
\label{traklka}
\end{equation}
% = &\min\limits_{\bm{\beta}^T \bm{\Lambda} \bm{\beta}=\bm{I}}Tr\left( (\bm{V}\bm\beta+\tilde{\bm{V}}\bm\gamma)^T\bm{K}(\bm{L}_w + \lambda \bm{L}) \bm{K}(\bm{V}\bm\beta+\tilde{\bm{V}}\bm\gamma) \right) \\
Note that $\bm{\Lambda}$ is invertible, so the solution $\bm\beta$ is the eigenvectors corresponding to $\bm{\Lambda}^T\bm{V}^T(\bm{L}_w+\lambda \bm{L})\bm{V\Lambda}$.
To enable the solution in the real domain, we can make 
\begin{equation}
	\bm\beta=\bm{\Lambda}^{-\frac{1}{2}}\bm\omega
  \label{beta}
\end{equation}
and the problem becomes
\begin{equation}
\label{obj_9}
\min\limits_{\bm{\omega}^T\bm{\omega}=\bm{I}} Tr(\bm\omega^T\bm{\Lambda}^{\frac{1}{2}}\bm{V}^T(\bm{L}_w + \lambda \bm{L}) \bm{V\Lambda}^{\frac{1}{2}}\bm\omega)
  %\label{obj_5}
\end{equation}
The optimal solution $\bm\omega$ is the eigenvectors of $\bm{\Lambda}^{\frac{1}{2}}\bm{V}^T(\bm{L}_w+ \lambda \bm{L}) \bm{V\Lambda}^{\frac{1}{2}}$ with respect to its eigenvalues in an ascending order. We summarize the entire procedure in Algorithm \ref{alg:1} to learn the new features. For any test data $\bm{x}^\prime$, its $l$-th new feature is obtained by $\sum\limits_{i=1}^n\alpha_{il} \bm{k}(\bm{x}^\prime,\bm{x}_i)$, $1\le l\le r$.
\LinesNumbered
\begin{algorithm}[!tb]
\caption{Local Discriminant Structure Uncovering.}
\label{alg:1}
\begin{algorithmic}[1]
\REQUIRE ~~\\
The training data matrix $\bm{X}=[\bm{X}_l,\bm{X}_u]\in\mathbb{R}^{d\times n}$\\
The classification assignment matrix $\bm{Y}=[\bm{Y}_l,\bm{Y}_u]\in\mathbb{R}^{n\times c}$
\ENSURE ~~\\
Feature transformation matrix $\bm a \in\mathbb{R}^{n\times r}$
\STATE Compute kernel matrix $\bm{K}$;\\
\STATE Compute graph $\bm{L}_w$ and $\bm{L}$ according to \eqref{Lw} and \eqref{LRGA_L};\\
\STATE Compute $\bm{V}$ and $\bm{\Lambda}$ by performing eigen-decomposition on $\bm{K}$ according to \eqref{K}; \\
\STATE Compute $\bm\omega$ by performing eigen-decomposition according to \eqref{obj_9};\\
\STATE Compute $\bm\beta$ according to \eqref{beta};\\
\STATE Obtain $\bm a$ according to \eqref{alpha};\\
\end{algorithmic}
\end{algorithm}
\section{Experiments}
In this section, we will briefly introduce the datasets and the compared methods which are used in the experiments. Afterwards, experimental results are evaluated and analyzed. 
\subsection{Datasets and Compared Methods}
To evaluate our algorithm, we have conducted extensive experiments and compared with a number of approaches on five datasets:
\begin{itemize}
\item \textbf{COIL-20 \cite{murase1995visual}}: It contains 1,440 gray-scale images of 20 objects (72 images per object) under various poses. The objects are rotated through 360 degrees and taken at the interval of 5 degrees.
\item \textbf{UMIST \cite{graham1998characterizing}}: The UMIST, which is also known as the Sheffield Face Database, consists of 564 images of 20 individuals. Each individual is shown in a variety of poses from profile to frontal views. 
\item \textbf{USPS \cite{hull1994adatabase}}: This dataset collects 9,298 images of handwritten digits (0-9) from envelops by the U.S. Postal Service. All images have been normalized to the same size of 16 $\times$ 16 pixels in grayscale. 
\item \textbf{Yale \cite{georghiades2001few}}: It consists of 2,414 frontal face images of 38 subjects. Different lighting conditions have been considered in this dataset. All images are reshaped into 24 $\times$ 24 pixels. 
\item \textbf{MIMIC II \cite{saeed2011multiparameter}}: It consists of 32,536 patient records in Intensive Care Unit (ICU) at the Beth Israel Deaconess Medical Center (BIDMC) collected from 2001 to 2008. We only extract medical notes from the database together with mortality information (if the patient has been expired at ICU for the first admission). A similar feature extraction pipeline used in \cite{kdd14} has been applied. Differently, Bag-of-Words model is used to encode multiple notes at different times for each patient. Empirically, we set the size of the dictionary as 500. Afterwards, we randomly select 1,000 adult patients of which are positive and negative evenly. 
\end{itemize}
For the first four datasets (COIL20, UMIST, USPS and YaleB), we just simply use their pixels as the input features and the typical RBF kernel.
For MIMIC II dataset, we use a 500 dimensional Bag-of-Words representation and a $\chi^2$ kernel. There are a number of kernel functions that have been invented so far. In this paper, our focus is to demonstration the effectiveness of our proposed algorithm rather than making comparisons between different kernels regarding classification performance. 

To evaluate our proposed algorithm, we choose several methods as the baseline:
\begin{itemize}
\item \textbf{Kernel Discriminant Analysis (KDA)} \cite{scholkopft1999fisher}: As one of the representative feature dimensionality reduction methods, KDA aims to project data into a direction on which class centers are far from each other while data samples of the same class are close to each other after feature mapping. We use a speed-up version implemented in \cite{cai11srkda}.
\item \textbf{Kernel Principal Component Analysis (KPCA)} \cite{scholkopf1998nonlinear}: Compared to KDA, KPCA reduces feature dimensionality by transforming data into a new coordinate system where the top n greatest variances of data correspondingly lie on the first n coordinates in the new subspace. We test different dimensionality reduction scenarios by KPCA across all the datasets. The best classification performance results are reported.
\item \textbf{Kernel Semi-supervised Discriminant Analysis (KSDA)} \cite{cai2007semi}: SDA aims to solve the problem of scarcity of label information when performing discriminant analysis. To utilize unlabeled samples, a graph Laplacian is built to approximate the local geometry of the data manifold where both the labeled and unlabeled data reside. Kernel SDA (KSDA) is used in the experiments.
\item \textbf{Kernel Semi-supervised Local Fisher discriminant analysis (KSELF)} \cite{sugiyama2010semi}: SELF which leverages supervised Local Fisher Discriminant Analysis and unsupervised Principal Component Analysis, is a linear semi-supervised dimensionality reduction method which makes feature analysis effective when only a small number of labeled samples are available. In the experiments, we use its non-linear extension termed as KSELF.
\item \textbf{Kernel Locality Preserving Projections (KLPP)} \cite{he2004locality}: KLPP is an unsupervised manifold learning method which preserves the local structure of samples, i.e. neighborhood relationship, in the original feature space as well as in the new projected space.   
\item \textbf{Co-regularized Ensemble for Feature Selection (EnFS)} \cite{han2013co}: This method employs a co-regularized framework in which a joint $\ell_{2,1}$-norm of multiple feature selection matrices can alleviate the over-fitting problem when the number of labeled data is small. Furthermore, a subset of feature that is more distinctive can be uncovered by removing irrelevant or noisy features. 
\end{itemize}
For all the methods, corresponding parameters are tuned in the same range of $\{10^{-4}$, $10^{-3}$, $10^{-2}$, $1$, $10^2$, $10^3$, $10^4\}$. Support Vector Machine (SVM) with a linear kernel has been applied as a classifier evaluating the classification performance of each method. The SVM parameter that controls the trade-off between the margin and the size of the slack variables is also tuned in the aforementioned range. The detailed dataset partition is followed by the convention of the semi-supervised learning approaches. Specifically, the training set contains both labeled and unlabeled data, and the testing set is not available during the training phrase. $c$ is denoted as the number of classes for each dataset. In the training dataset, we randomly sample a number of labeled data per class (1, 3, 5, and 10) as different class settings. Therefore the numbers of labeled training data are $1\times c$, $3\times c$, $5\times c$, and $10\times c$ in the different class settings, while the remaining training data are treated as unlabeled. Particularly, we also investigate conditions in which more labeled information is available by using $25\times c$, $50\times c$, $75\times c$, and $100\times c$ on MIMIC II dataset. We repeat the experiments five times using the data partitions mentioned above, and report the average results. Because we focus on classification performance, we use Mean Average Precision (MAP) as the effectiveness measure in the comparisons.
\subsection{Evaluation}
We have made comparisons among all the methods mentioned above over five datasets under different label settings. Generally speaking, our algorithm consistently achieves better classification performance under different settings than all the counterparts. Specifically, from Tab. \ref{tab_overall}, we can observe that our approach outperforms all supervised and unsupervised methods, including KDA, KPCA and EnFS, when labeled data are quite few, i.e. $1\times c$. For instance, a number of relatively large margins can be observed on UMIST and USPS. Compared to the semi-supervised methods (KSDA, KSELF) and the unsupervised manifold learning method (KLPP), our method still has superior performance over all the datasets under different settings. In the conditions where more labels are available, it is observed that our method still performs better than all the compared approaches. However, the difference margins between our method and the other approaches are quite limited. For example, our method, on the MIMIC II dataset, performs quite similarly to EnFS and KSELF. 

\setlength\tabcolsep{2pt}
\begin{table*}[!tb]\scriptsize
\caption{Performance comparison (Mean Average Precision $\pm$ STD) across all datasets with a linear SVM classifier.}
\centering
\begin{tabular}{|l|c|c|c|c|c|c|c|c|}
\hline
Dataset& Settings& KDA & KPCA & KSDA & KSELF & KLPP & EnFS & Ours\\
\hline
\multirow{4}{*}{COIL-20}  	& $1\times c$  & 0.728$\pm$0.026&0.699$\pm$0.026&0.777$\pm$0.016&0.702$\pm$0.032&0.780$\pm$0.020&0.702$\pm$0.012&\textbf{0.818$\pm$0.001}\\
							& $3\times c$  & 0.832$\pm$0.013&0.833$\pm$0.018&0.840$\pm$0.020&0.819$\pm$0.014&0.822$\pm$0.007&0.825$\pm$0.025&\textbf{0.850$\pm$0.005}\\
							& $5\times c$  & 0.888$\pm$0.015&0.881$\pm$0.020&0.878$\pm$0.022&0.877$\pm$0.014&0.876$\pm$0.015&0.880$\pm$0.013&\textbf{0.897$\pm$0.002} \\
							& $10\times c$ & 0.948$\pm$0.009&0.949$\pm$0.012&0.926$\pm$0.015&0.931$\pm$0.012&0.932$\pm$0.007&0.935$\pm$0.012&\textbf{0.957$\pm$0.001} \\
\hline
\multirow{4}{*}{UMIST}  	& $1\times c$  &0.574$\pm$0.019&0.558$\pm$0.021&0.642$\pm$0.026&0.573$\pm$0.018&0.580$\pm$0.020&0.558$\pm$0.017&\textbf{0.654$\pm$0.010}\\
							& $3\times c$  &0.882$\pm$0.033&0.851$\pm$0.035&0.860$\pm$0.034&0.832$\pm$0.044&0.812$\pm$0.043&0.810$\pm$0.037&\textbf{0.889$\pm$0.001}\\
							& $5\times c$  &0.960$\pm$0.013&0.942$\pm$0.019&0.945$\pm$0.020&0.937$\pm$0.019&0.913$\pm$0.021&0.909$\pm$0.031&\textbf{0.963$\pm$0.001}\\
							& $10\times c$ &0.995$\pm$0.004&0.989$\pm$0.003&0.987$\pm$0.008&0.990$\pm$0.004&0.967$\pm$0.006&0.9779$\pm$0.01&\textbf{0.996$\pm$0.002}\\
\hline
\multirow{4}{*}{USPS}  		& $1\times c$  &0.536$\pm$0.079&0.464$\pm$0.056&0.653$\pm$0.088&0.351$\pm$0.023&0.789$\pm$0.056&0.523$\pm$0.053&\textbf{0.801$\pm$0.009}\\
							& $3\times c$  &0.727$\pm$0.017&0.720$\pm$0.019&0.811$\pm$0.015&0.625$\pm$0.073&0.905$\pm$0.012&0.715$\pm$0.027&\textbf{0.936$\pm$0.008}\\
							& $5\times c$  &0.795$\pm$0.024&0.788$\pm$0.032&0.865$\pm$0.031&0.732$\pm$0.028&0.939$\pm$0.017&0.788$\pm$0.024&\textbf{0.948$\pm$0.007}\\
							& $10\times c$ &0.868$\pm$0.018&0.860$\pm$0.007&0.911$\pm$0.007&0.837$\pm$0.011&0.957$\pm$0.005&0.864$\pm$0.016&\textbf{0.964$\pm$0.003}\\
\hline
\multirow{4}{*}{YaleB}  	& $1\times c$  &0.359$\pm$0.013&0.358$\pm$0.015&0.245$\pm$0.007&0.220$\pm$0.011&0.354$\pm$0.028&0.216$\pm$0.014&\textbf{0.463$\pm$0.004}\\
							& $3\times c$  &0.872$\pm$0.023&0.761$\pm$0.033&0.555$\pm$0.034&0.652$\pm$0.033&0.572$\pm$0.033&0.567$\pm$0.083&\textbf{0.891$\pm$0.001}\\
							& $5\times c$  &0.951$\pm$0.003&0.907$\pm$0.006&0.769$\pm$0.021&0.856$\pm$0.008&0.717$\pm$0.020&0.723$\pm$0.039&\textbf{0.964$\pm$0.002}\\
							& $10\times c$ &0.981$\pm$0.003&0.978$\pm$0.005&0.948$\pm$0.005&0.963$\pm$0.011&0.826$\pm$0.018&0.863$\pm$0.032&\textbf{0.992$\pm$0.007}\\
\hline
\multirow{4}{*}{MIMIC II}  	& $25\times c$ &0.683$\pm$0.029&0.675$\pm$0.028&0.666$\pm$0.051&0.699$\pm$0.017&0.683$\pm$0.037&0.696$\pm$0.028&\textbf{0.700$\pm$0.027}\\
						    & $50\times c$ &0.727$\pm$0.017&0.722$\pm$0.026&0.705$\pm$0.035&0.733$\pm$0.014&0.712$\pm$0.004&0.740$\pm$0.008&\textbf{0.741$\pm$0.018}\\
							& $75\times c$ &0.741$\pm$0.012&0.749$\pm$0.027&0.727$\pm$0.024&0.751$\pm$0.010&0.737$\pm$0.020&0.754$\pm$0.028&\textbf{0.755$\pm$0.015}\\
							& $100\times c$&0.754$\pm$0.025&0.757$\pm$0.021&0.759$\pm$0.019&0.757$\pm$0.028&0.751$\pm$0.020&0.762$\pm$0.023&\textbf{0.766$\pm$0.017}\\
\hline
\end{tabular}
\label{tab_overall}
\end{table*}
\begin{figure*}[!b]
  \centering
  \subfigure[COIL20]{
    \label{COIL20_dim} %% label for first subfigure
    \includegraphics[width = .45\linewidth]
    {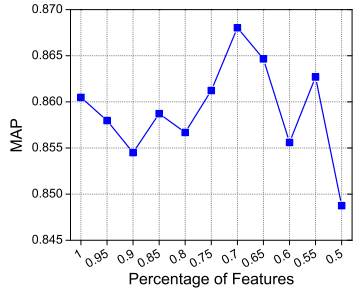}}\vspace{-0mm}\hspace{0mm}    
  \subfigure[UMIST]{
    \label{UMIST_dim} %% label for first subfigure
    \includegraphics[width = .45\linewidth]
    {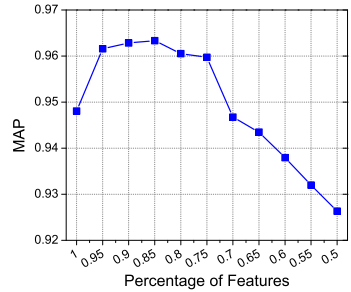}}\vspace{-0mm}\hspace{0mm}
   \subfigure[USPS]{
    \label{USPS_dim} %% label for first subfigure
    \includegraphics[width = .45\linewidth]
    {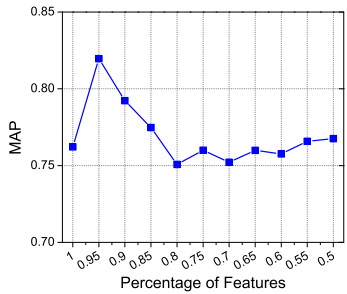}}\vspace{-0mm}\hspace{0mm}
   \subfigure[YaleB]{
    \label{YaleB_dim} %% label for first subfigure
    \includegraphics[width = .45\linewidth]
    {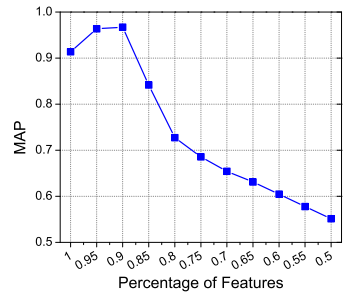}}\vspace{-0mm}\hspace{0mm}
\caption{Performance variations w.r.t dimensionality reduction parameter $r$.}
\label{fig_dimensionality} %% label for entire figure
\end{figure*}
Apart from the overall classification performance comparisons in Tab. \ref{tab_overall}, we also studied the effects of parameters used in our method. Due to the page limit, we only studied the sensitivity of parameters on four image datasets. Note that some of the parameters are fixed for demonstration in the following experiments. Thus, the performance results are not as good as the ones in Tab. \ref{tab_overall} in which all parameters are tuned to achieve the best performance. In the first round, we test parameter sensitivity for $\theta$ and $k$ which are both required when constructing the new graph Laplacian. We test $\theta$ in the range of $\{10^{-4}, 10^{-3}, 10^{-2}, 1, 10^2, 10^3, 10^4\}$ and $k$ in the range of $[1,3,5,10,15,20]$. We find that the system is not sensitive to $\theta$ and the optimal $k$ should be 3 or 5.

\begin{figure*}[!b]
  \centering
  \hspace{-5mm}
  \subfigure[COIL20]{
    \label{COIL20_lda} %% label for first subfigure
    \includegraphics[width = .45\linewidth]
    {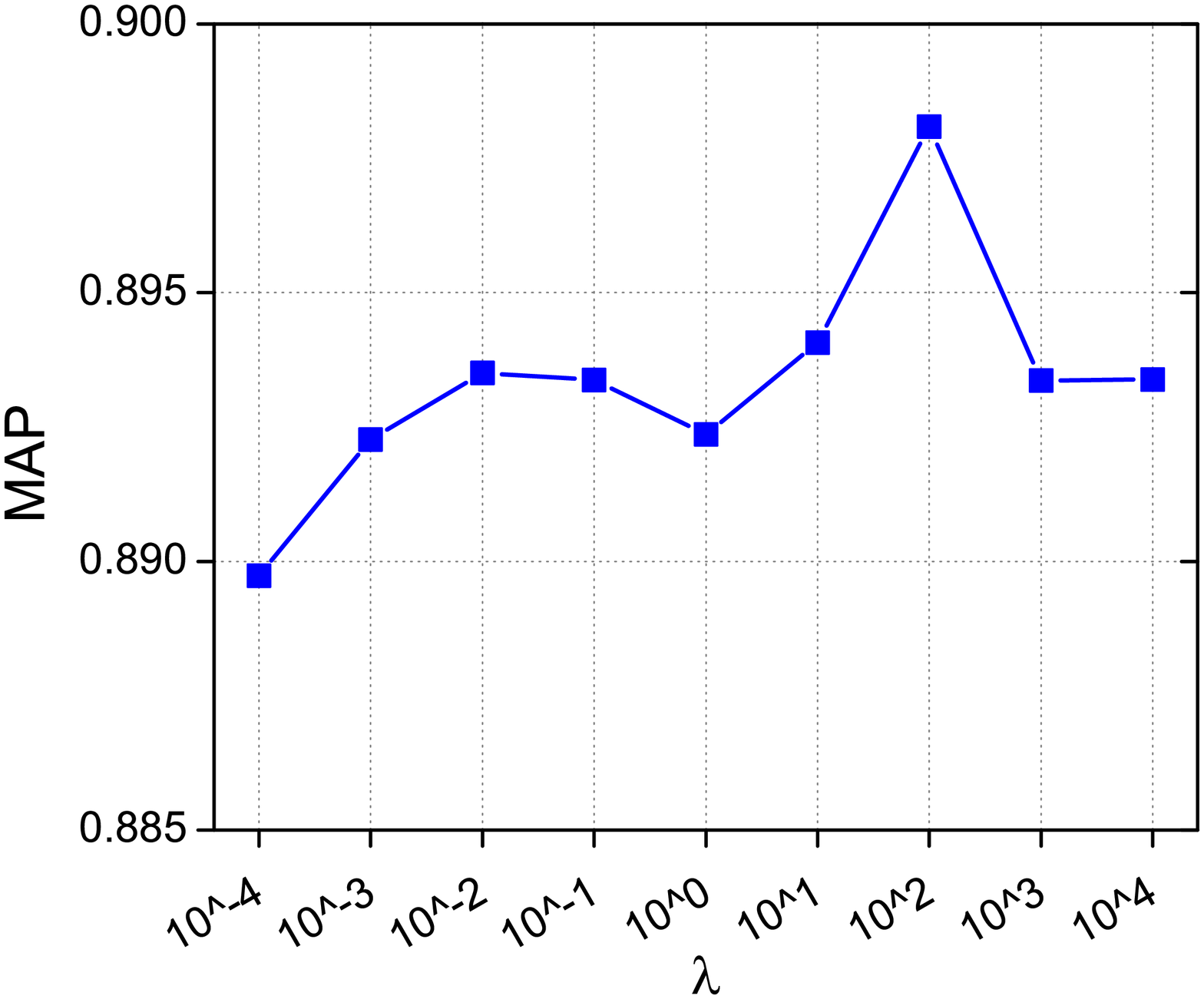}}
  \subfigure[UMIST]{
    \label{UMIST_lda} %% label for first subfigure
    \includegraphics[width = .45\linewidth]
    {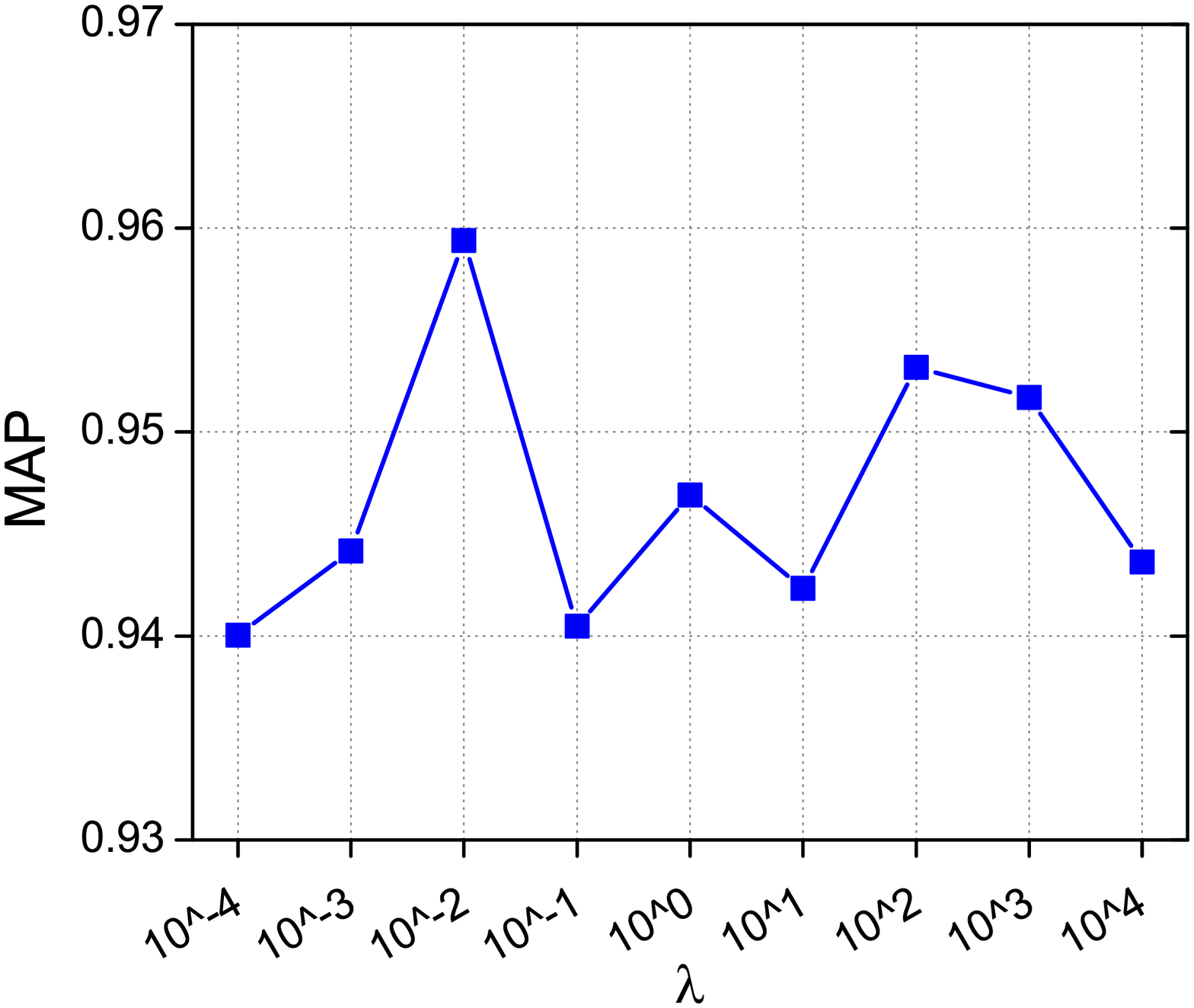}}\vspace{-0mm}\hspace{0mm}   
  \subfigure[USPS]{
    \label{USPS_lda} %% label for first subfigure
    \includegraphics[width = .45\linewidth]
    {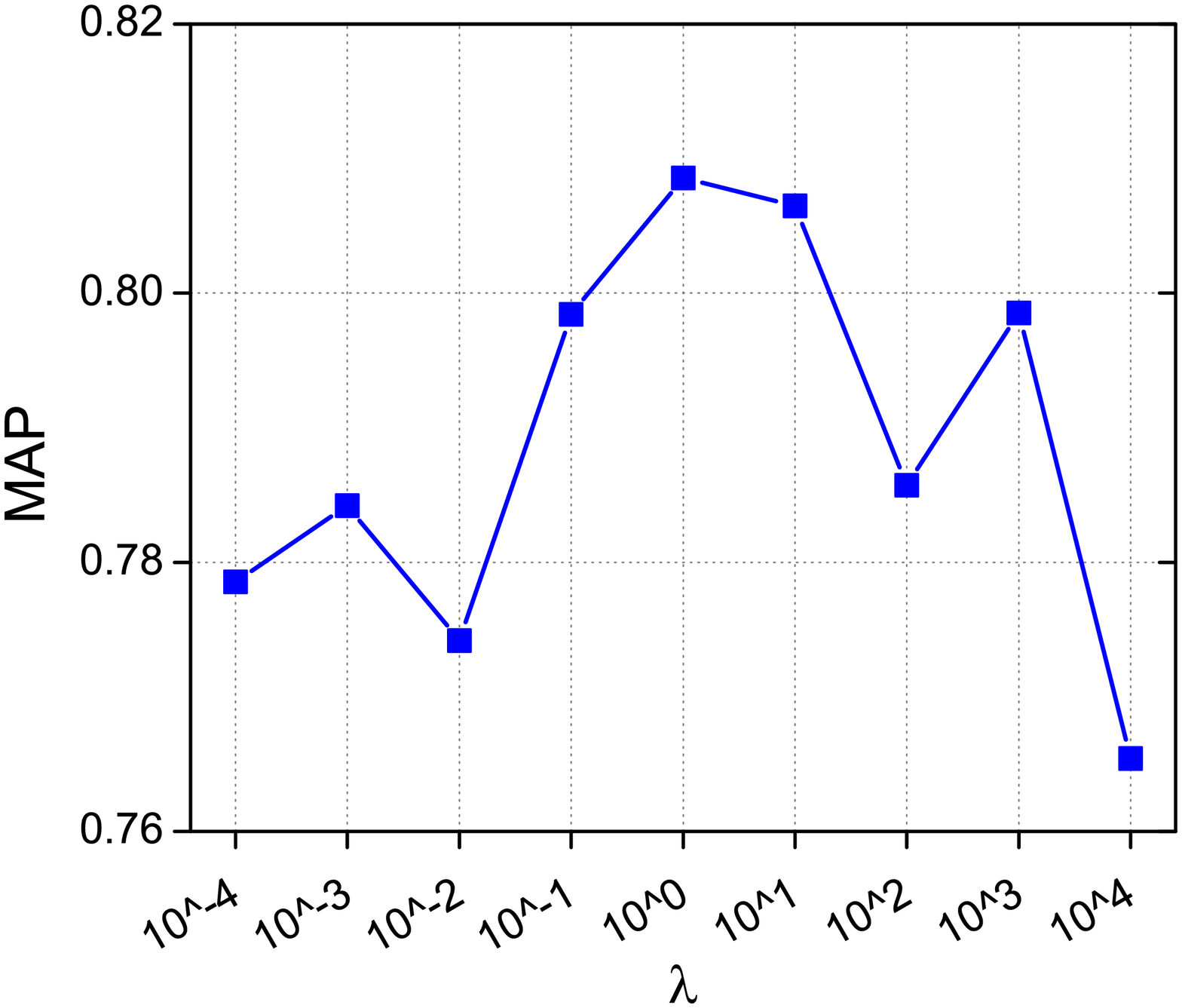}}\vspace{-0mm}\hspace{0mm}
  \subfigure[YaleB]{
    \label{YaleB_lda} %% label for first subfigure
    \includegraphics[width = .45\linewidth]
    {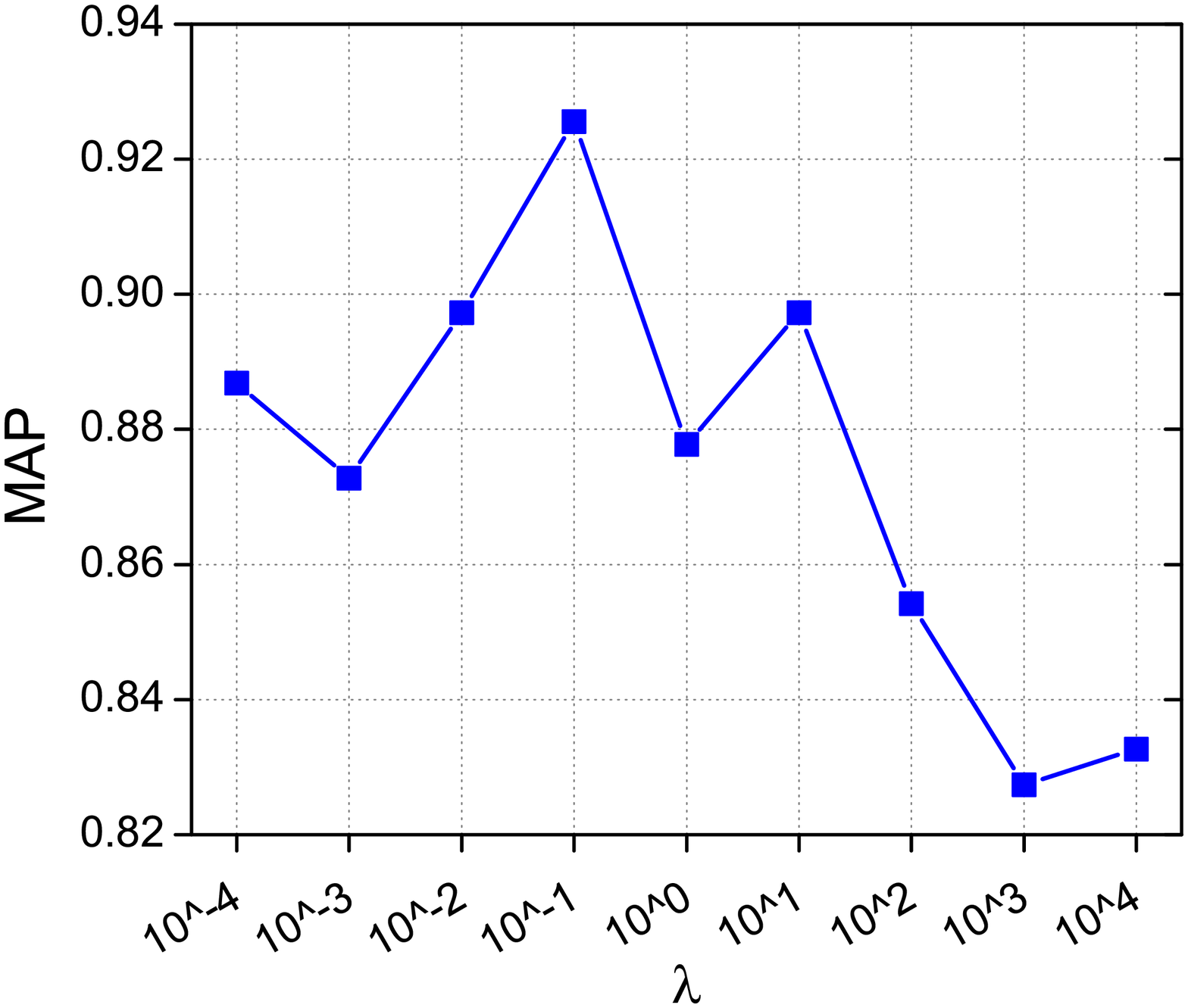}}\vspace{-0mm}\hspace{0mm}
\caption{Performance variations w.r.t regularization parameter $\lambda$.}
\label{fig_lda} %% label for entire figure
\end{figure*}
In the second round, we firstly fix the aforementioned two parameters of graph Laplacian, $\theta=1$ and $k=3$, respectively. Moreover, the regularization parameter, $\lambda$, in \eqref{obj_9} is set to 1. We plot classification performance changes over four datasets when dimensionality reduction parameter, $r$, varies. Our aim is to understand how the new features with reduced dimensionality impact performance by fixing all the parameters except the dimension of the inferred feature. The results show that we do not have to use full dimensional features. For example, in Fig. \ref{COIL20_dim}, the performance scores 86.04\% when 100\% learned features are preserved on COIL20. It then peaks at 86.67\% when 70\% are preserved. For each dataset, a further improvement has been observed after reducing dimensionality of the new features. This improvement might be because irrelevant features and noises are removed by our method after feature mapping in the higher dimensional space.  

In the last experiment, all three parameters are fixed ($\theta=1$, $k=3$ and $r=1$). We compare the variations of classification performance when changing the regularization parameter, $\lambda$, which leverages local manifold structures in the framework. Note that there is no contribution from local discriminant structure analysis when $\lambda$ is close to zero. From Fig. \ref{fig_lda}, the performance on each dataset is a relatively lower value when little local manifold information has been considered. With the variations of $\lambda$, for each dataset, the performance varies and scores the best when the weight of the graph Laplacian is increased to a certain amount which is obviously greater than $0$. For example, in Fig. \ref{UMIST_lda}, the performance starts around $94\%$ and almost peaks at $96\%$ when $\lambda=0.01$, with a nearly $2\%$ improvement. This result confirms that our algorithm successfully incorporates local manifold information into the feature analysis procedure.    
\section{Conclusion}
In this paper, we have proposed a semi-supervised feature analysis method. Specifically, our method enforces data from the same class to become closer to each other in a high-dimensional space after feature mapping. In order to take both local discriminant information and manifold structure into account, a local discriminant model has been applied to the local clique of each data point. Our method successfully learns both labeled and unlabeled data via leveraging the new graph Laplacian that holds local discriminant information. It has proven that our method effectively learns features when the number of labeled data points is quite small. 
\section*{Acknowledgement}
This work was supported by Australian Research Council Discovery Project. The project number is DP140100104. Any opinions, findings, and conclusions or recommendations expressed in this material are those of the author(s) and do not necessarily reflect the views of the Australian Research Council. 
\bibliographystyle{splncs03}
%\begin{thebibliography}{}
%\bibitem{nene1996columbia}
%%Sameer A Nene, Shree K Nayar, Hiroshi Murase and others
%Columbia object image library (COIL-20)  
%Technical Report CUCS-005-96, 1996
%\end{thebibliography}
\bibliography{mybibfile}
\end{document}